\documentclass{article}
\usepackage{authblk}
\usepackage[T1]{fontenc}
\usepackage{multirow}
\usepackage[table,xcdraw]{xcolor}

\usepackage{framed,enumitem}

\usepackage{graphicx}
\title{Safe Deep RL for Intraoperative Planning of Pedicle Screw Placement 
}
\author[1, 2, 4]{Yunke Ao}
\author[1]{Hooman Esfandiari}
\author[1]{Fabio Carrillo}
\author[2, 4]{Yarden As}
\author[1]{Mazda Farshad}
\author[3, 4]{Benjamin F. Grewe}
\author[2, 4]{Andreas Krause}
\author[1, 4]{Philipp Fürnstahl}
\affil[1]{ROCS, Balgrist University Hospital, University of Zurich, Switzerland}
\affil[2]{Department of Computer Science, ETH Zurich, Switzerland}
\affil[3]{Department of Information Technology and Electrical Engineering, ETH Zurich, Switzerland}
\affil[4]{ETH AI Center, ETH Zurich, Switzerland}

\providecommand{\keywords}[1]
{
  \small	
  \textbf{\textit{Keywords---}} #1
}
\date{}                     
\begin{document}
\maketitle              
\begin{abstract}
Spinal fusion surgery requires highly accurate implantation of pedicle screw implants, which must be conducted in critical proximity to vital structures with a limited view of anatomy. Robotic surgery systems have been proposed to improve placement accuracy, however, state-of-the-art systems suffer from the limitations of open-loop approaches, as they follow traditional concepts of preoperative planning and intraoperative registration, without real-time recalculation of the surgical plan. In this paper, we propose an intraoperative planning approach for robotic spine surgery that leverages real-time observation for drill path planning based on Safe Deep Reinforcement Learning (DRL). The main contributions of our method are (1) the capability to guarantee safe actions by introducing an uncertainty-aware distance-based safety filter; and (2) 
the ability to compensate for incomplete intraoperative anatomical information, by encoding a-priori knowledge of anatomical structures with a network pre-trained on high-fidelity anatomical models.
Planning quality was assessed by quantitative comparison with the gold standard (GS). In experiments with 5 models derived from real magnetic resonance imaging (MRI) data, our approach was capable of achieving $90\%$ bone penetration with respect to the GS while satisfying safety requirements, even under observation and motion uncertainty. Up to our knowledge, our approach is the first safe DRL approach focusing on orthopedic surgeries.

\end{abstract}
%
%
%
\section{Introduction} 

Pedicle screws play a critical role in spinal instrumentation to achieve primary stability in fusion procedures.
%
Given the limited visibility of the underlying anatomy during surgery, the surgical execution remains challenging due to the proximity to vital structures (e.g., spinal cord), particularly if performed conventionally.
%
Therefore, various 
robotic technologies have been developed to improve the accuracy of pedicle screw placement (PSP)~\cite{abe2011novel,lieberman2006bone,lieberman2020robotic}. 
However, they rely heavily on a pre-defined preoperative plan to position the end-effector~\cite{farber2021robotics,d2019robotic}, thereby failing to consider larger intraoperative changes to the surgical plans.

To address this limitation, one current research direction is the integration of additional intraoperative data (e.g., ultrasound (US)~\cite{li2022automatic}, fluoroscopy~\cite{jecklin_x23dintraoperative_2022} or RGB-depth (RGB-D) cameras~\cite{liebmann2021spinedepth}) to improve the real-time decision-making capabilities of the robotic surgery system.
%
However, these intraoperative data are challenging to be utilized by current planning approaches for PSP~\cite{zhang20183d,qi2022automatic}, as these approaches are based on preoperative imaging with all information of relevant anatomies available.
In contrast, real-time intraoperative data only provides partial, noisy observations.
For instance, RGB-D cameras can only capture the exposed surface of the vertebra~\cite{liebmann2021spinedepth} and US imaging provides only limited field-of-view and penetration depth~\cite{li2023robot}.

For planning based on noisy and partial observation, recent robotic research beyond the domain of orthopedic surgery have demonstrated the effectiveness of deep reinforcement learning (DRL)~\cite{lee2020learning,miki2022learning,kalashnikov2018qt,scheikl2022sim}.
%
However, so far their application in clinical settings has been limited, primarily due to the safety concern that high-risk actions can still be taken by learned policies~\cite{pore2021safe}.
Various safe RL frameworks have been proposed to improve the safety performance of DRL~\cite{gu2022review}.
But their application in orthopedic surgery has not yet been investigated.

In this work, we aim at developing a DRL method for intraoperative planning of robotic drilling for PSP, which leverages partial observations for continuous updates while guaranteeing safe operation.
To achieve automatic planning with no harm to crucial anatomical structures, we propose a novel safe DRL agent that learns to incorporate distance-based safety filters inspired by~\cite{selim2022safe,shao2021reachability,alshiekh2018safe,as2022constrained}.
While traditional systems based on preoperative planning have difficulty to keep surgery plans up to date, our approach can continuously update based on partial and noisy observations, which makes it suitable for intraoperative modalities. 
%


\section{Method} 

\begin{figure}[t]
\centering
\includegraphics[width=0.95\textwidth]{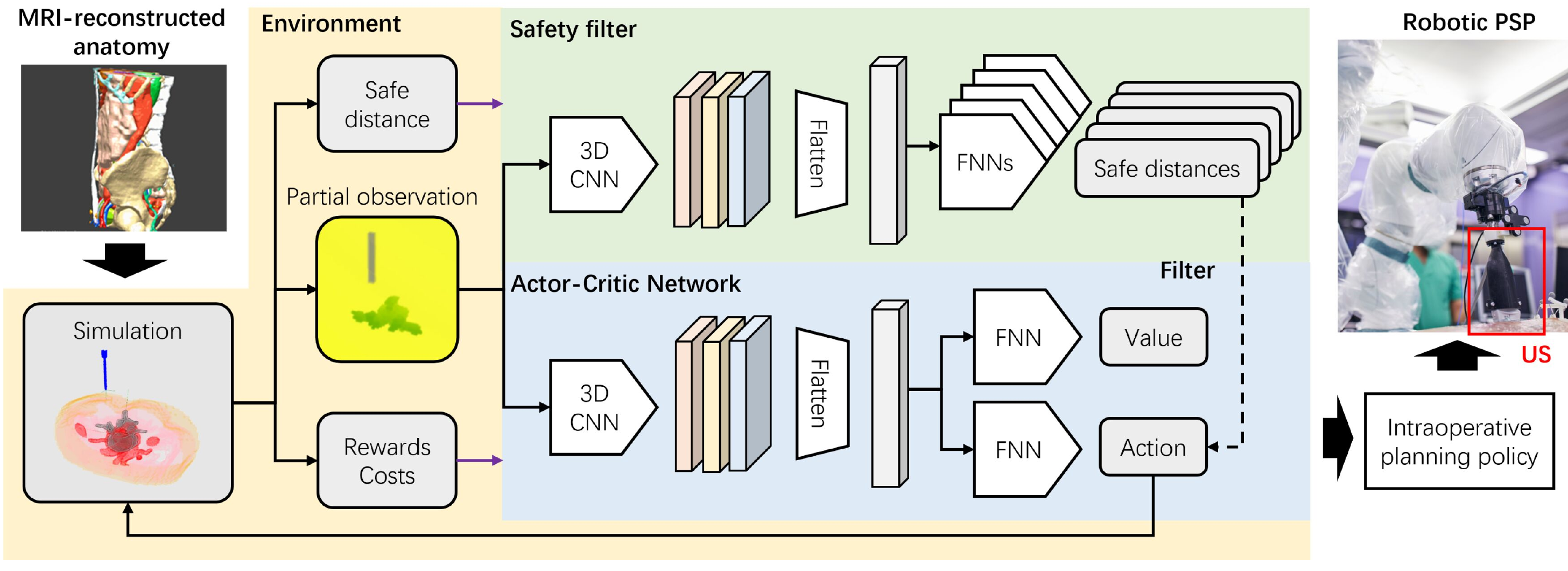}
\caption{Overview of our method. 
%
%
%
The training environment utilized MRI-reconstructed anatomical model to simulate intraoperative partial observation and generate rewards, costs and safe distance to train a DRL agent.
The DRL agent processed the partial observation with 3D CNN, then output a motion command for the drill.
A safety-filter was also trained and only utilized during policy deployment to filter the output. 
%
%
} 
\label{fig:overview}
\end{figure}

\vspace{-0.6em}
An overview of the proposed framework for training a DRL agent is shown in Figure~\ref{fig:overview}. 
Robotic drilling for PSP was modeled as a constrained Markov decision process (CMDP).
Simulations were constructed based on detailed anatomical structures surrounding the target vertebra.
Partial observation mimicking intraoperative data was extracted as input to the agent, which consisted of a distance-based safety filter and an actor-critic network.
%
%
The trained agent is foreseen to be used for intraoperative planning in the context of robotic PSP~\cite{li2022automatic}. 

\subsection{Task Definition for Intraoperative Planning of PSP}
\label{sec:prob} 
We pursue a minimal invasive approach where percutaneous robotic drilling is performed through the anatomical structures for PSP~\cite{huntsman2020navigated}.
In our setting, the patient is positioned prone for a better dorsal reach and US scan is performed before starting the procedure. 
 The dorsal surface of the target vertebra is reconstructed from US scan according to existing approaches such as~\cite{ding2003automatic,hacihaliloglu2013statistical,li2022automatic}. 
Based on this input, the agent should output the real-time motion command for the robotic drill.
The reconstructed dorsal surface is updated after each motion step, and new commands are generated by the agent according to new inputs.
%
%
%
While repeating this sequence, the drill is expected to first reach the entry point of the skin, then perforate through the subcutaneous tissue, fascia, muscles, cortical and cancellous bone until reaching the desired penetration length. 
Bone breach or soft tissue injury should be avoided and the drilled path should enable anchoring the screw stably into the bone.

%
%

%
%

\subsection{Simulation Environment}
\label{sec:sim}
\begin{figure}[]
\centering
\includegraphics[width=0.95\textwidth]{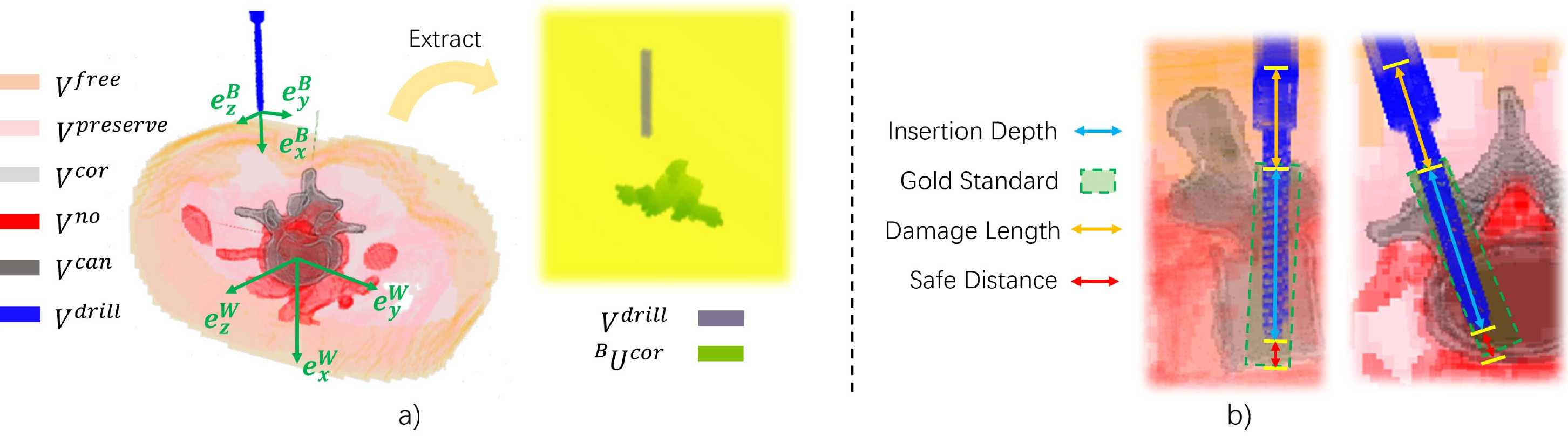}
\caption{
Simulation environment for PSP. 
a) Anatomical groups and partial observation;
b) Terms defined for training the agent.
%
GS was represented by a cylinder constructed from the cancellous entry point, exit point, insertion direction and pedicle width.
} \label{fig:simulation}
\end{figure}

%
A simplified simulation was constructed for this intraoperative planning scenario.
It utilized a detailed 3D representation of all relevant anatomies surrounding the target vertebra.
 These anatomies were categorized into 5 groups: tissue that can be dissected or damaged (skin), tissues on which the damage should be minimized (muscles), cortical bone, cancellous bone, and no go regions (organs, nerves).
The 3D volumes for these anatomical groups were represented mathematically as $V^{free}$, $V^{preserve}$, $V^{cor}$, $V^{can}$ and $V^{no}$, respectively, and were shown in Figure~\ref{fig:simulation}.
The drill was modeled as cylinders with sufficient lengths and diameters $d$ equal to those of the pedicle screws, which is denoted as $V^{drill}$.
Thus, safe status was defined as |$V^{drill}\cap V^{no}|=0$ and no breaking through $V^{can}\cup V^{cor}$ by $V^{drill}$.
This was identified by a function $safe(V^{drill}, V^{can}, V^{cor}, V^{no})$
, which returns 1 if the current status is safe and 0 otherwise.
The goal of anchoring the screw stably could be simplified as maximizing $|V^{can}\cap V^{drill}|$.
This is reasonable because maximizing cancellous penetration will result in the screw being inserted into the central area of the pedicle and surrounded by cortex.

In our simulation, the world frame $W$ and the drill frame $B$ were defined  with axes shown in Figure~\ref{fig:simulation}. 
The 3D motion of the drill was described as $({^B\delta} {x_t}, {^B\delta} {y_t}, {^B\delta}{z_t}, \beta_t, \gamma_t)$, corresponding to $XYZ$ translations and rotations around $YZ$ axes in the drill frame.
Rotation around $e^B_x$ was neglected, as the drill was considered symmetric around $e^B_x$.
At each step, the agent could only select one axis to either move or rotate the drill in positive or negative direction with a predefined scale.
%
Gaussian and uniform motion noises were added to simulate motion uncertainties from limited robot accuracy and patient movement, respectively.  
Motion constraints were imposed by setting small side-wise motion scales when the drill inserted into soft tissues or bones.
The detailed step-wise motion scales and noises levels are shown in Table~\ref{tab:motion}.

The partial observation $^BU^{cor}$ was simulated from the ground truth model of the cortical bone by extracting its dorsal surface, as is illustrated in Figure~\ref{fig:simulation}. 
To simulate potential intraoperative reconstruction errors~\cite{li2023robot}, Gaussian noise with 1 mm standard deviation (STD) and uniform noise between -2 mm and 2 mm were introduced to positions of surface points.

\begin{table}[]
\caption{Motion parameters in our simulation.
"Soft", "Cortical" and "Cancellous" means the drill tip is inside the soft tissues, cortical and cancellous bones, respectively.
The noise value denotes both the STD for Gaussian noise and range of uniform noise.
}
\footnotesize
\centering
\scalebox{0.85}{
\begin{tabular}{cc|cc|cc|cc|cc}
\hline
\multicolumn{2}{c|}{\multirow{2}{*}{\textbf{Motions}}}                           & \multicolumn{2}{c|}{\textbf{Outside}} & \multicolumn{2}{c|}{\textbf{Soft}} & \multicolumn{2}{c|}{\textbf{Cortical}} & \multicolumn{2}{c}{\textbf{Cancellous}} \\ \cline{3-10} 
\multicolumn{2}{c|}{}                                                            & \textbf{    scale    }    & \textbf{    noise    }    & \textbf{    scale    }        & \textbf{    noise    }       & \textbf{scale}       & \textbf{noise}       & \textbf{scale}        & \textbf{noise}        \\ \hline
\multicolumn{1}{c|}{\multirow{3}{*}{\textbf{Translation {[}mm{]}}  }} & \textbf{ x } & 5.00              & 0.15              & 4.00                  & 0.40                 & 1.00                 & 0.10                 & 2.00                  & 0.20                  \\
\multicolumn{1}{c|}{}                                               & \textbf{ y } & 2.00              & 0.15              & 0.40                  & 0.10                 & 0                    & 0                    & 0                     & 0                     \\
\multicolumn{1}{c|}{}                                               & \textbf{ z } & 2.00              & 0.15              & 0.40                  & 0.10                 & 0                    & 0                    & 0                     & 0                     \\ \hline
\multicolumn{1}{c|}{\multirow{2}{*}{\textbf{Rotation {[}rad{]}}}}   & \textbf{ y } & 0.050             & 0.002             & 0.010                 & 0.001                & 0                    & 0                    & 0                     & 0                     \\
\multicolumn{1}{c|}{}                                               & \textbf{ z } & 0.020             & 0.002             & 0.004                 & 0.001                & 0                    & 0                    & 0                     & 0                     \\ \hline
\end{tabular}
}
\label{tab:motion}
\end{table}

\subsection{Constrained Markov Decision Process Modeling}\label{sec:CMDP}

Robotic drilling was modeled as a CMDP described by a tuple $(S, A, R, C, T, \rho)$, representing state space, action space, reward function, cost function, transition model and initial state distribution~\cite{altman1999constrained}.
The objective of CMDP is to maximize the expected total rewards while satisfying constraints on expected total costs.

At each time step $t$, the state $s_t$ representing the partial observation was defined as a $100\times100\times20$ 3D volume with a 5 mm grid size depicting the current position of the dorsal surface with respect to the drill.
As is shown in Figure~\ref{fig:simulation}, each voxel $^Bp_i$ was labeled as 1 or 2 depending on whether $^Bp_i \in {^BV^{drill}_t}$ or $^Bp_i \in {^BU^{cor}_t}$ respectively. 
Other voxels are labeled as 0.
The action $a_t$ was an integer between 1 and 11, with 1-10 mapped to the aforementioned 5 dimensional drill motion $({^B\delta} {x_t}, {^B\delta} {y_t}, {^B\delta}{z_t}, \beta_t, \gamma_t)$, and 11 corresponding to "no movement".
%
%
 
Our step-wise cost was defined as $c_t= safe(V^{drill}_t, V^{can}, V^{cor}, V^{no}) - 1$.
We aimed at constraining total cost to 0 to prevent unsafe behaviors.
To encourage the learned policy to satisfy this constraint, we include $c_t$ in the reward design.
The resulting safety-aware reward $r$ has four terms $r = w_1 r^{bone} + w_2 r^{n-dam} + w_3 r^{n-enter} + w_4 r^{recover}$, which are modeled as follows:

\begin{itemize}[align=left]
    \item [$r^{bone}$] We define cancellous penetration depth as: $b_t = \frac{4}{\pi d^2}|V^{drill}_t\cap V^{can}_t|$, 
    and the reward as: $r^{bone} = b_t - b_{t-1}$.
    The agent will receive this reward only if the drill is on the correct side (left or right pedicle).
    \item [$r^{n-dam}$]
    Total damaged region is derived from $V^{dam}_t = V^{dam}_{t-1}\cup (V^{drill}_t\cap V^{little}_t)$, and damage length is defined as $m^{dam}_t =  \frac{4}{\pi d^2}|V^{dam}_{t}|$.
    Then $r_t^{n-dam}$ is modeled as negative additional damage length: $m^{dam}_{t-1} - m^{dam}_t$.
    \item [$r^{n-enter}$] In cases when the agent moves from a safe state into an unsafe region ($c_{t-1}=0$ and $c_t=1$), a penalty was applied based on the penetration rewards gathered so far: $r^{n-enter} = - (1 - c_{t-1}) \cdot c_t \cdot \sum_{\tau=0}^t r^{bone}_\tau$. 
    This could motivate the agent to prioritize safety. 
    \item [$r^{recover}$] This reward incentivizes the agent to recover as quickly as possible once inside an unsafe region: $r^{recover} = c_t\cdot (I\{a_t=1\} - I\{a_t=0\})$. 
    $I$ is the indicator of whether the action is translating forward (1) or backward (0).
\end{itemize}

The weights of the rewards were tuned to strike a balance between meeting clinical requirement and enabling exploration.
We selected $w_3=0.4$ and $w_1=1.0$ such that a significant penalty was imposed for violating the constraint. 
We set $w_2=0.01$ for prioritizing minimally invasiveness after optimizing the first 2 objectives, followed by $w_4 = 2$ for encouraging fast recovery.

\subsection{Safe RL with distance-based safety filter}\label{sec:safe_rl}
%
As stated in related literature, it is difficult to ensure that learned policies using standard DRL will not exhibit any unsafe actions.
Therefore, we designed a safe DRL agent that learns to better satisfy the constraints for total costs within our predefined simulation and CMDP model.
As illustrated in Figure~\ref{fig:overview}, our approach consisted of a novel safety filter and an actor-critic network, both of which utilized a feature extractor based on 3D Convolutional Neural Network (CNN) layers.
The actor-critic network generated actions and value predictions, and was trained using the Proximal Policy Optimization (PPO) algorithm~\cite{schulman2017proximal}.
%
Network structures and hyperparameters are detailed in the supplementary material.

The safety filter used 5 fully-connected neural network (FNN) heads to predict the safe distance $l^{safe}$,
defined as the minimum translation along the drill direction to reach an unsafe state (Figure~\ref{fig:simulation}).
This is because the safety is violated almost always by the forward moving action due to motion constraints.
5 heads were trained with different samples of experiences using the mean squared error (MSE) to the ground truth values as the loss function.
Epistemic uncertainties were then estimated from: ${\sigma^{l}_t}^2 = \frac{1}{5} \sum_{i=1}^5 (\hat{l}^{safe}_{t, i} - \hat{m}^l_t)^2$, where $\hat{m}^l_t = \frac{1}{5} \sum_{i=1}^5 \hat{l}^{safe}_{t, i}$.
During training, the safety filter was not utilized for decision-making to ensure a proper rate of exploration, but it was implemented to filter the real-time actions during policy deployment.
With the pessimism similar to~\cite{as2022constrained,selim2022safe}, the agent was forced to take the "backward moving" action if $\hat{m}^l_t - \lambda \sigma^{l}_t < 0$, where $\lambda$ is an adjustable constant to trade off between bone penetration and safety.
%
%
Otherwise, it would select the action from the actor network.

\subsection{Experimental Setup}\label{sec:exp}

\textbf{Dataset and training environment}
We used ITIS virtual population V3 and V4 models~\cite{gosselin2014development} to create our simulation environment and evaluate our method.
These models contained fully segmented anatomies generated from real magnetic resonance imaging (MRI) data. 
From this dataset, we chose 5 adult models with different body mass indexes (BMIs), genders and ages named: Duke, Ella, Fats, Jeduk and Glenn.
We developed simulation environments only for their lumbar vertebrae (L1-L5, 25 environments in total).
Gold standard (GS) trajectory, pedicle width (PW) and cancellous entry and exit points for each vertebra were annotated by expert surgical planners of our in-house 3D planning center of the clinic. 
%
%
We selected the screw diameter (SD) mainly to be $70\%$ of the PW to balance the training difficulty with the surgical requirements mentioned in~\cite{solitro2019currently}.

\textbf{Evaluation protocol}
We conducted two groups of experiments: Individual group and cross validation group.
In the former, we trained and tested our policy with the same human model.
This form of evaluation was implemented to support the surgical scenario in which MRI / computational tomography (CT) scans are available prior to the surgery, from which fully segmented anatomies could be obtained to construct the simulation environment for training the agent.
%
In the cross validation group, we evaluated the generalization ability of our method over anatomy variations.
We used each human model to test the performance of a policy trained with the 4 other human models.
We randomly scaled the anatomies from $0.9-1.1$ along each axis to augment the training dataset.

\textbf{Evaluation metrics}
Each policy was assessed based on safe rate, cancellous penetration depth, damage length, and deviation from GS.
%
Averaged performance of each policy was computed from 200 trajectories generated by the policy with random initial drill poses above the anatomy and randomly selected vertebrae from L1-L5.
The safe rate measured the proportion of trajectories that do not encounter any unsafe states.
%
For GS, cancellous penetration and damage length were computed from the extended trajectories that continue along the back direction until it exits the human body.
The angle between directions and distance between entry points were computed to evaluate the difference between our trajectories and GS plannings.


\section{Results and discussion} 

\begin{figure}[t]
\centering
\includegraphics[width=\textwidth]{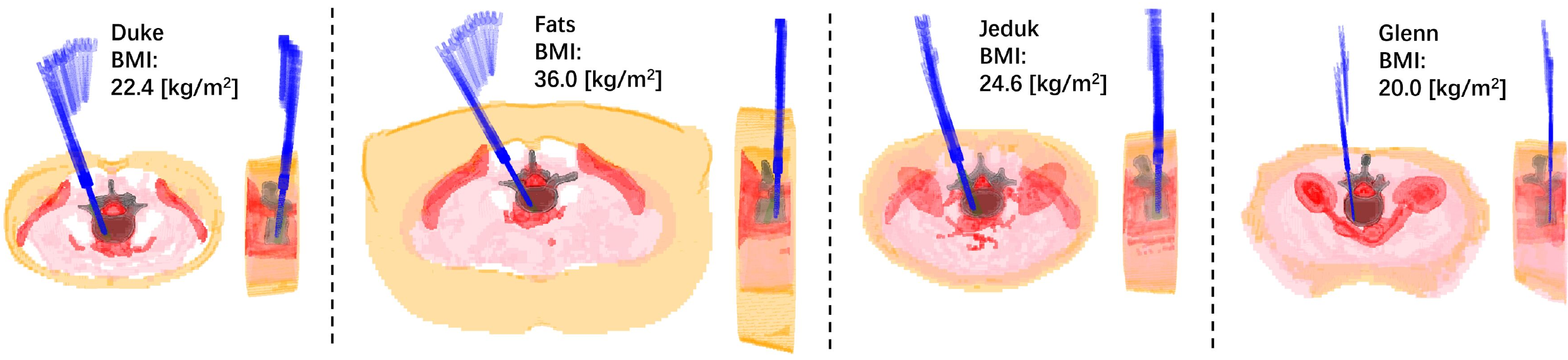}
\caption{Forward and side views of example trajectories planned by our policy in the individual group. 
The vertebrae from different levels and different human models with various BMIs are selected to demonstrate the effectiveness of our method.  
}
\label{fig:exp_traj}
\end{figure}
\begin{figure}[t]
\centering
\includegraphics[width=\textwidth]{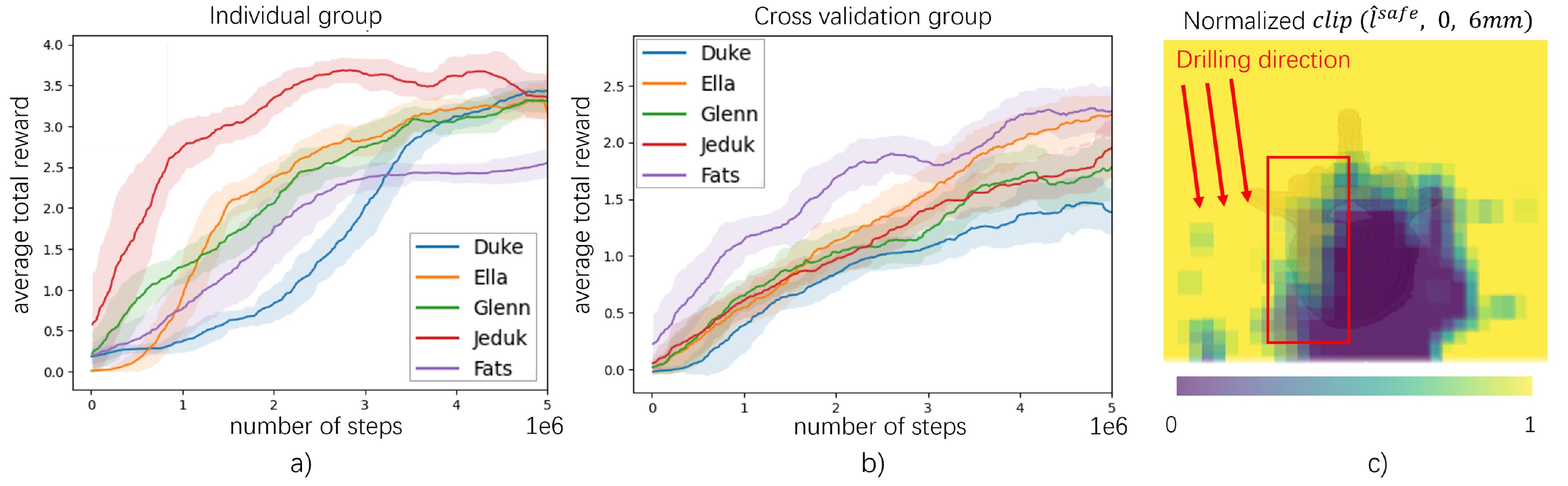}
\caption{Experiment results.
a) Learning curves of the individual group;
b) Learning curves of the cross validation group; 
c) Prediction of $\hat{l}^{safe}$ along the shown drill direction when the drill tip is at each position. 
Horizontal and lateral axes are $x$ and $y$ positions.
%
%
}
\label{fig:learning_curve}
\end{figure}

\textbf{Individual group}
%
In our experiments, each agent was trained for more than 5M steps (around 10 hours), as illustrated in Figure~\ref{fig:learning_curve}.
%
For this group, each agent satisfied safety constraints in all 200 cases, and the average penetration depth was more than $90\%$ of that of GS, with a slight increase in the average damage to the restricted area (33$\%$), as is shown in Table~\ref{tab:evaluation}. 
This increase was due to the motion noise embedded in the simulation, and could be minimized by integrating additional safety mechanisms into the robot in a real-life setup.
%
%
%
Additionally, Figure~\ref{fig:exp_traj} demonstrates that our resulting trajectories have desired penetration depths.
The average angle between our trajectories and GS was 0.146 rad, and the average distance between entry points was 4.71 mm.
These results indicate that if the complete anatomy data of the patient is available before the surgery, even with partial observations and motion uncertainties, our approach is able to plan safe motions with close performance to the GS.

\textbf{Cross validation group}
%
As is shown in Table~\ref{tab:evaluation}, generalization performance varies for different human models, particularly for outliers such as Fats shown in Figure~\ref{fig:exp_traj}. 
On average, the angle and entry point difference from the GS were 0.15 rad and 4.43 mm, respectively.
 The safety rates of policies without the safety filter was lower than those of the individual group due to anatomical variations.
However, after incorporating the safety filter, the safety performance was largely improved to surpass $90\%$, albeit at the expense of reducing cancellous penetration.
This indicates that the agent becomes more cautious in uncertain situations, which is desirable in surgery where safety is most important.

\textbf{Learning of relevant bone shape from reward and safe distance}
Our approach could compensate for partial data which do not contain structures such as the anterior bone part or soft tissues. 
This is because during training, the agent received rewards and safe distance based on the complete bone shape, and relevant information was stored in the value network and safety filters.
This is illustrated in Figure~\ref{fig:learning_curve}, where the distribution of predicted safe distance captures the depth of the vertebra (purple shadow) along the drilling direction.

\textbf{Limitations}
With only partial observation, generalization over variation of anatomies was inherently difficult, as the unknown shape of a new vertebra was hard to predicted based solely on its dorsal surface.
This difficulty was compounded by the small size of the human model dataset used.
Also, our environment must be further developed in the future to support realistic soft tissue deformation.
Finally, there were still sim-to-real gaps between our simulation and the real scenarios. 
%
Thus, simulating realistic input from US or RGBD efficiently and overcoming sim-to-real gap would be subject of future work. 

\begin{table}[t]
\centering
\caption{Numerical testing results. 
$w/ f$ and $w/o f$  represent with and without safety filter, respectively. 
"Ideal" denotes the gold standards.
Penetration depths of some policies are greater than gold standards because they violate the safe constraints.
}
\footnotesize
\scalebox{0.85}{
\begin{tabular}{c|c|cc|ccc|ccc}
\hline
\multirow{2}{*}{\textbf{Groups}} & \multirow{2}{*}{\textbf{Humans}} & \multicolumn{2}{c|}{\textbf{Safe rate {[}\%{]}}} & \multicolumn{3}{c|}{\textbf{Penetration Depth {[}mm{]}}}  & \multicolumn{3}{c}{\textbf{Damage Length{[}mm{]}}}          \\ \cline{3-10} 
                                 &                                  & \textbf{w/o f}     & \textbf{w/ f}     & \textbf{   w/o f   } & \textbf{   w/ f   } & \textbf{ideal} & \textbf{  w/o f  } & \textbf{  w/ f  } & \textbf{ideal} \\ \hline
\multirow{5}{*}{\textbf{Individual}}      & Duke                             & 89.0                    & 100.0                  & 51.9                & 45.0                 & 53.0             & 59.0                  & 50.9              & 37.3           \\
                                 & Glenn                            & 100.0                   & 100.0                  & 49.4                & 49.3               & 50.1           & 106.0                 & 107.0             & 90.4           \\
                                 & Ella                             & 37.5                    & 100.0                  & 51.7                & 45.2               & 48.2           & 68.1                & 53.3              & 31.6           \\
                                 & Jeduk                            & 96.0                    & 100.0                  & 53.6                & 49.4               & 55.5           & 131.0                 & 121.0             & 85.9           \\
                                 & Fats                             & 99.5                    & 100.0                  & 44.9                & 44.4               & 49.5           & 36.8                & 36.0              & 30.5           \\ \hline
\multirow{5}{*}{\textbf{Cross Val}}      & Duke                             & 71.5                    & 98.5                   & 52.4                & 36.6               & 52.7           & 62.4                & 47.6              & 34.1           \\
                                 & Glenn                            & 79.5                    & 90.0                   & 51.4                & 30.9               & 50.4           & 116.0                 & 99.5              & 89.6           \\
                                 & Ella                             & 27.0                    & 92.0                   & 46.1                & 32.8               & 48.2           & 68.7                & 40.5              & 32.2           \\
                                 & Jeduk                            & 92.0                    & 95.5                   & 56.2                & 44.3               & 55.5           & 124.0                 & 114.0             & 86.0             \\
                                 & Fats                             & 24.5                    & 100.0                  & 36.3                & 23.7               & 49.5           & 60.9                & 29.5              & 31.3           \\ \hline
\end{tabular}
}
\label{tab:evaluation}
\end{table}

%


\section{Conclusion}
In this study, we have presented a safe DRL agent for intraoperative planning of orthopedic surgery of the spine.
Up to our knowledge, our approach is the first safe DRL approach focusing on orthopedic surgeries. 
A particular distance-based safety filter was designed, and intraoperative data was simulated  from virtual anatomical structures to train the DRL agent for real-time intraoperative planning.
By addressing the challenges of partial observation and safety, our method could provide the basis for intraoperative decision-making and robotic surgery with higher level of automation.

\bibliographystyle{splncs04}
\bibliography{samplepaper}
%




\end{document}